%% file: root.tex
\PassOptionsToPackage{dvipsnames,table}{xcolor}

\documentclass[letterpaper, 10 pt, conference]{ieeeconf}  %

\IEEEoverridecommandlockouts                              %

\usepackage{graphics} %
\usepackage{epsfig} %
\usepackage{mathptmx} %
\usepackage{times} %
\usepackage{amsmath} %
\usepackage{amssymb}  %
\usepackage{lib}
\usepackage{alias}
\usepackage{xspace}
\usepackage{xcolor}
\usepackage[addedmarkup=uline]{changes}
\usepackage{booktabs}
\usepackage{blindtext}
\usepackage{colortbl}
\usepackage{multirow}
\usepackage[labelfont=bf]{caption}
\usepackage{soul}
\usepackage{cuted}
\let\labelindent\relax
\usepackage{enumitem}

\definecolor{linkpink}{HTML}{D81B60}      %
\usepackage{hyperref}
\hypersetup{
  colorlinks = true,
  linkcolor  = black,    %
  citecolor  = black,    %
  urlcolor   = linkpink  %
}

\definechangesauthor[name=A, color=red]{a}

\title{\LARGE \bf
\changedSinceFirstRAL{Precise Mobile Manipulation of Small Everyday Objects}
}

\author{Arjun Gupta \quad Rishik Sathua \quad Saurabh Gupta \\
University of Illinois at Urbana-Champaign \\
\href{https://arjung128.github.io/svm/}{arjung128.github.io/svm}}

\newif\ifcolorize
\colorizefalse

\newcommand{\changedSinceFirstRAL}[1]{%
  \ifcolorize
    {\color{blue}#1}%
  \else
    #1%
  \fi
}

\begin{document}

\twocolumn[{%
\renewcommand\twocolumn[1][]{#1}%
\maketitle
\vspace{-20pt}
\begin{center}
\insertW{1.0}{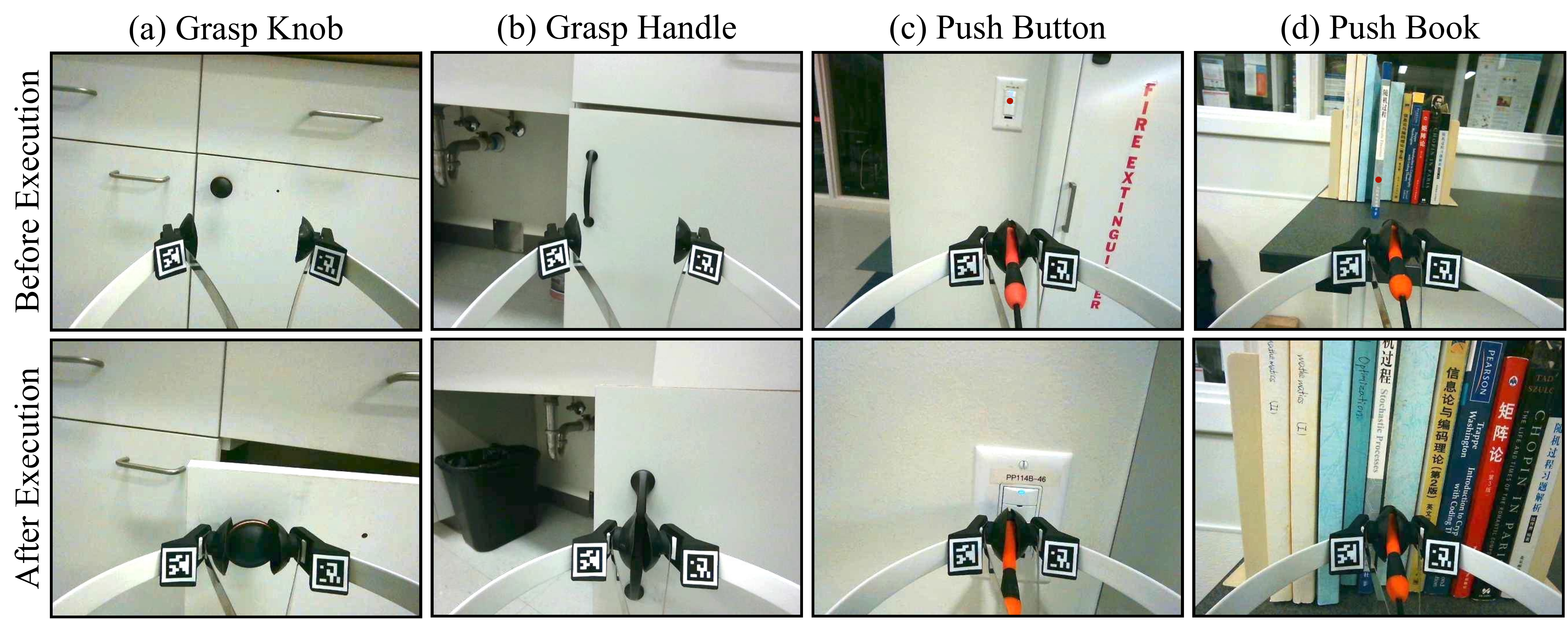}
\captionof{figure}{Many everyday mobile manipulation tasks require reaching a precise interaction site before executing a motion primitive, \eg precise reaching of a knob / handle to pull open a cupboard in (a) and (b), or precisely reaching a user-indicated button / book before pushing it in (c) and (d) (shown via the {\color{red} red dot}). Open loop execution is unable to meet the high-precision needed for these tasks. In this paper, we develop Servoing with Vision Models (\name), a training-free framework that closes the loop to enable a commodity mobile manipulator to tackle these tasks.}
\figlabel{teaser}
\end{center}
}]
\bibliographystyle{IEEEtran}

\thispagestyle{empty}
\pagestyle{empty}

\begin{abstract}

Many everyday mobile manipulation tasks require precise interaction with small objects, such as grasping a knob to open a cabinet or pressing a light switch. In this paper, we develop Servoing with Vision Models (\name), a closed-loop framework that enables a mobile manipulator to tackle such precise tasks involving the manipulation of small objects. 
\changedSinceFirstRAL{\name uses state-of-the-art vision foundation models to generate 3D targets for visual servoing to enable diverse tasks in novel environments.
Naively doing so fails because of occlusion by the end-effector. \name mitigates this using vision models that out-paint the end-effector thereby significantly enhancing target localization.}
We demonstrate that aided by out-painting methods, open-vocabulary object detectors can serve as a drop-in module for \name to seek semantic targets (e.g. knobs) and point tracking methods can help \name reliably pursue interaction sites indicated by user clicks. 
\changedSinceFirstRAL{We conduct a large-scale evaluation spanning experiments in 10 novel environments across 6 buildings including 72 different object instances. \name obtains a 71\% zero-shot success rate on manipulating unseen objects in novel environments in the real world, outperforming an open-loop control method by an absolute 42\% and an imitation learning baseline trained on 1000+ demonstrations also by an absolute success rate of 50\%.}

\end{abstract}

\input{1-intro-v3}

\input{2-related}

\input{3-method}

\input{4-experiments}

\input{5-discussion}

\input{6-acknowledgements}

\bibliography{biblioShort.bib, references.bib}

\end{document}

%% file: 1-intro-v3.tex
\section{Introduction}
\seclabel{intro}

\begin{figure*}
\insertW{1.0}{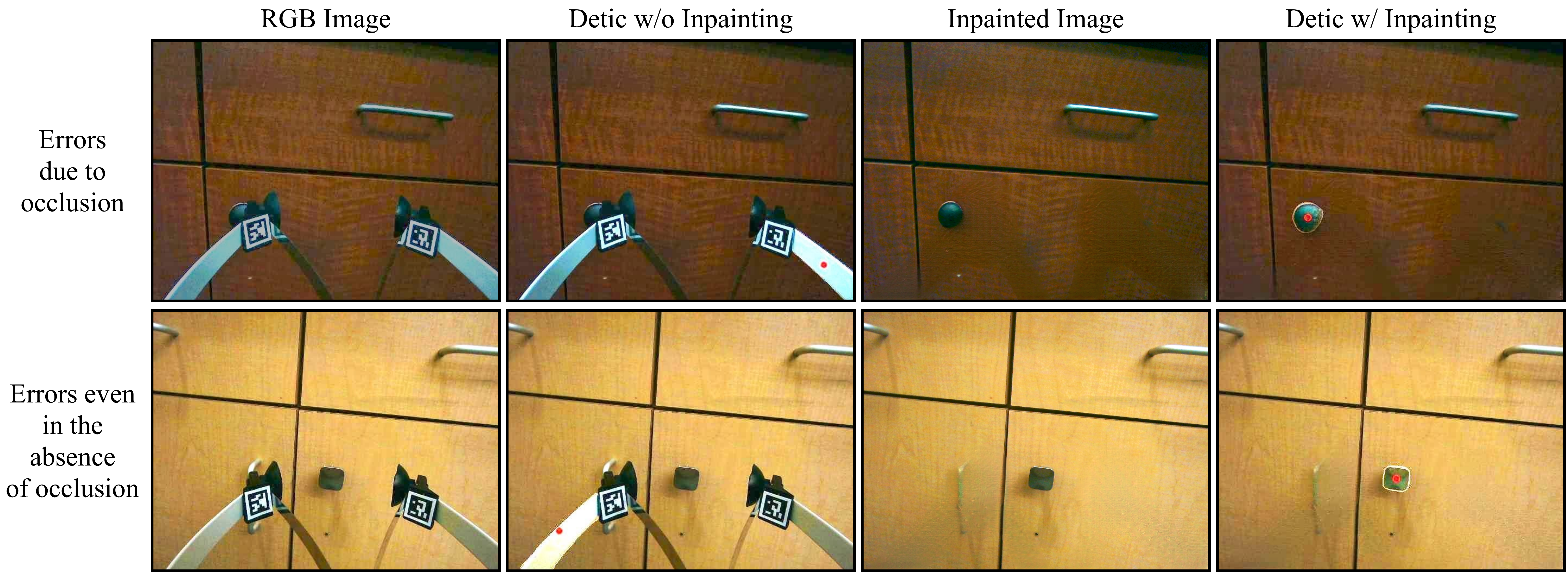}
\caption{\changedSinceFirstRAL{When using off-the-shelf detectors on wrist camera data, knob detections (indicated by the {\color{red} red point} in the second column) are incorrect. Errors stem from occlusion of the knob due to the end-effector ({\bf top}) and due to the presence of the end-effector (out-of-distribution object) even when the knob is unoccluded ({\bf bottom}). Out-painting the end-effector ({\bf right two columns}) fixes this.}}
\figlabel{occlusion}
\end{figure*}

Mobile manipulators hold the promise of performing a wide range of useful tasks in
our 
everyday environments.
However, a major obstacle to realizing this vision lies in the lack of precise mobile manipulation capabilities of current systems.
Many real world
tasks require precise interaction with small objects, such as grasping a knob to pull open a cabinet or pressing a light switch, 
where even a small deviation can cause failure.
A mobile manipulator's mobility makes these tasks even more challenging.
For example, small errors during navigation, say on a 
thick
carpeted surface, can easily exceed the tight tolerance required for precise tasks, and a non-holonomic base may limit precise repositioning.
Furthermore, mobility means that mobile manipulators have to manipulate in varied locations under diverse lighting and unnatural viewpoints 
(\eg looking top-down at a drawer from very close by), demanding significantly broader generalization than stationary manipulators confined to 
a fixed environment.
As a result, developing mobile manipulators 
capable of performing 
precise tasks while generalizing to diverse environments remains an open problem.

Many precise mobile manipulation tasks involve {\it stylized} interactions with small objects: reaching a precise interaction site before executing a simple motion. 
The difficulty in these tasks lies in reaching the accurate pre-task pose around the small target object 
whereas the subsequent motion is easy to execute.
For getting to the accurate pre-task pose, open-loop execution using a sense-plan-act
paradigm does not work because inaccuracies in perception and control prevent
correct engagement with the small interaction site. 
This necessitates a closed-loop approach. Imitation learning would be a natural choice, but as our experiments will show, policies trained with imitation fail to achieve sufficient precision and generalization even when trained on 1000+ real world demonstrations.
So, how can we achieve precise mobile manipulation of small everyday objects?

Visual servoing with an eye-in-hand camera is an effective
technique to close the loop to precisely pursue
targets~\cite{chaumette2016visual}
without requiring large-scale task-specific training for broad generalization, 
unlike closed-loop imitation learning.
However, vanilla visual servoing makes
strong assumptions, \eg requiring known 3D objects or target images, which are
not available in our in-the-wild setting. Our proposed approach, Servoing with 
Vision Models or \name, marries together
visual servoing with modern vision foundation models to mitigate these limitations. 
This leads to an effective system which is able to operate in a closed-loop
manner, and at the same time is versatile enough to operate in novel
environments on previously unseen objects. 

\changedSinceFirstRAL{\name leverages vision foundation models in two ways.}
First, we use them to
specify targets for the visual servoing module. This alleviates the need for
known 3D objects or target images. We experiment with two ways to specify
targets: a) semantic categories, and b) points of interaction. For objects that
have a well-known semantic category (\eg drawer knobs or cabinet handles), we
use an open-world object detector (\eg Detic~\cite{zhou2022detecting}) to
continuously detect the target during visual servoing. However, not all mobile manipulation interaction sites, \eg the different buttons on a microwave, correspond to a semantic
category. We tackle such cases by using point trackers (\eg CoTracker~\cite{karaev2023cotracker}) to continuously track a user-indicated interaction site (\eg a single user click in the first image, specifying which button on the microwave to push) during visual servoing. Thus, the use of vision foundation models takes care of the target specification problem in visual servoing.

\begin{figure*}
\insertW{1.0}{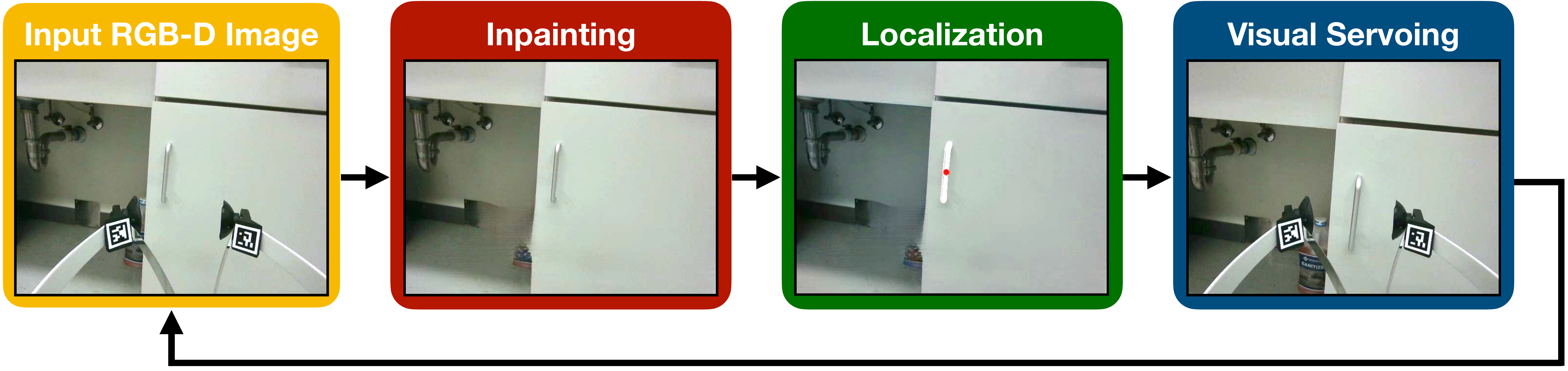}
\caption{{\bf Servoing with Vision Models (\name)} is a  framework for precise reaching for mobile manipulators. Starting from an input RGB-D wrist camera image with a target specified either via a semantic label (\eg handle) or a user-clicked point on the image, \name outputs whole-body control commands to convey the end-effector to the target location by closing the loop with visual feedback. \name first paints out the end-effector using a video outpainting model, uses vision foundation models to continuously detect the target object (or track the desired target point) to compute 3D servoing targets, which are passed to a servo to obtain whole-body control commands (see \secref{method}).}
\figlabel{overview}
\end{figure*}

One problem, however, still remains. The use of visual servoing with an eye-in-hand
camera for manipulation tasks suffers due
to occlusion of the environment by the manipulator. Such occlusion can be
particularly detrimental if it leads to out-of-distribution input to the
vision model that now starts producing erroneous predictions (see
second column of \figref{occlusion}). We mitigate this issue using yet another advance in
computer vision: video inpainting models~\cite{chang2023look,
zhou2023propainter, lee2019copy}. We out-paint the robot end-effector to obtain
a \textit{clean} view of the scene (see right two columns of \figref{occlusion}). This improves the
detection performance of the vision foundation model, leading to improved overall
success.

\changedSinceFirstRAL{We conduct experiments in 10 novel environments across 6 buildings to test \name on several real world tasks: grasping a knob to pull open
drawers / cupboards, grasping a handle to pull open a cabinet, pushing buttons, toggling switches, and pushing books into place on bookshelves. We obtain a 
71\% success rate on these challenging tasks, \textit{zero-shot on novel objects
in previously unseen environments}. As expected, \name performs much better than
open-loop control which only succeeds on 29\% of trials. SVM without end-effector 
out-painting only obtains a 62\% success rate, suggesting that %
the presence of
the end-effector does degrade performance and that our out-painting 
strategy is able to successfully mitigate it.}
Somewhat surprisingly, vision foundation models (Detic~\cite{zhou2022detecting}, CoTracker~\cite{karaev2023cotracker}, and ProPainter~\cite{zhou2023propainter} in our experiments)
perform quite well on out-of-distribution wrist camera images. 

Perhaps most interesting is the comparison of \name \vs imitation learning. 
As discussed, fine-grained, closed-loop tasks are precisely where one
would expect imitation learning to excel over a modular learning approach.
So it is natural to ask how well \name performs \vs imitation learning. We
conduct this comparison on the tasks of pulling knobs and handles for opening
articulated objects. Specifically, we compare to the recent Robot Utility Models
(RUM) work from Etukuru \etal~\cite{etukuru2024robot}. RUM is a closed-loop policy
trained on 1200
demonstrations for opening cabinets and 525 demonstrations for opening drawers
and thus serves as a very strong imitation learning comparison point. 
\changedSinceFirstRAL{Surprisingly, 
we find that \name outperforms RUM by an absolute success rate of
50\% across $30$ trials.}

To summarize, we develop \name, an approach to enable precise
mobile manipulation in the real world. This is made possible by
marrying together visual servoing with modern vision foundation models. Our experiments
reveal the effectiveness of our proposed approach over a number of strong baselines.
These results suggest that \name can serve
as an
effective alternative to imitation learning for 
generalizable and precise mobile manipulation of small everyday objects.

%% file: 2-related.tex
\section{Related Work}
\seclabel{related}

\textbf{Vision Foundation Models (Detection, Point Tracking, and Inpainting).}
Training on Internet-scale datasets~\cite{radford2021learning, achiam2023gpt,
kirillov2023segment} with large-capacity models~\cite{dosovitskiy2021image} has dramatically improved the generalization performance of
vision systems. This coupled with alignment of visual representations with ones
from language (\eg CLIP~\cite{radford2021learning}) has led to
effective open-vocabulary object detectors, \eg Detic~\cite{zhou2022detecting},
OVR-CNN~\cite{zareian2021open}.
Similar advances in diffusion-based generative
models~\cite{sohl2015deep, rombach2021high, ho2020denoising, song2020score} and
large-scale training have led to effective image generation models. These models have been leveraged for image and video
inpainting~\cite{zhou2023propainter, chang2023look}. Inpainting models have
also been used in robotics to mitigate domain gap between human and
robot data~\cite{bahl2022human, chang2023look}. 
Last, point-based tracking in videos is seeing renewed interest in recent
times~\cite{harley2022particle, karaev2023cotracker, doersch2023tapir}. 
Given a set of 2D points in the first frame, these
models are able to track them over a video. Use of machine learning makes
these new approaches more robust than earlier versions~\cite{shi1994good}. Forecasts of point tracks into the future have been
used as a intermediate representation for policy learning in
robotics~\cite{wen2023any, bharadhwaj2024track2act}. 

\textbf{Visual Servoing.}
Visual servoing (image-based,
pose-based, and hybrid approaches) outputs
control commands that convey the camera (and the attached manipulator) to a
desired location~\cite{corke1993visual,
chaumette2016visual, chaumette2006visual}.  Research has investigated use of
different features to compute distance between current and target images:
photometric distance~\cite{collewet2011photometric}, matching
histograms~\cite{bateux2016histograms}, features from pre-trained neural
networks~\cite{lee2017learning}, and has even trained neural networks to
directly predict the relative geometric transformation between
images~\cite{bateux2018training}. Visual servoing has been applied for 
manipulation~\cite{sadeghi2018sim2real}, 
navigation~\cite{sadeghi2019divis, li2021survey, qureshi2021rtvs}, 
1-shot visual imitation~\cite{argus2020flowcontrol} and for
seeking far away targets via intermediate view 
synthesis~\cite{crombez2021subsequent}. 

\changedSinceFirstRAL{\textbf{Precise Manipulation.}
Many recent works have tackled various \textit{precise} manipulation tasks \cite{tang2023industreal, luo2025precise, goyal2024rvt}. ForceSight \cite{collins2024forcesight} uses a neural network to predict visual-force goals for precise manipulation tasks on the Stretch. However, it trains a specialized affordance model on a small dataset of objects, in contrast to our approach of leveraging a vision foundation model, which subsequently leads to ForceSight's poor generalization (see supplementary video for more details). 
Most similar to our work is MOSART \cite{gupta2024opening}, which tackles opening of articulated objects using a modular system. Despite its strong generalization performance, this is an open-loop method, and as our experiments will show,
struggles on high-precision tasks. We introduce a novel methodology for closing-the-loop for such modular systems to enable precise manipulation.}

\textbf{Imitation Learning (IL).}
\changedSinceFirstRAL{IL~\cite{pomerleau1991efficient, schaal1996learning} is a
general tool for learning closed-loop manipulation policies. Recent research has made many advances: expressive policy architectures~\cite{chi2023diffusionpolicy, mahi2022behavior}, output representations~\cite{zhao2023learning}, and use of object-centric representations~\cite{zhu2022viola}.}
However, this generality comes with the
need for a large number of demonstrations for
generalization~\cite{khazatsky2024droid}. 
\changedSinceFirstRAL{IL has also been
successfully applied to eye-in-hand settings~\cite{chi2024universal, shafiullah2023bringing, etukuru2024robot}.}
Recent one/few-shot imitation learning
methods~\cite{valassakis2022demonstrate, wen2022you, vecerik2023robotap} %
leverage the structure of the task (getting to a bottleneck pose + motion
replay) to learn from a one / few demonstrations but are then restricted to
interacting with the object they were trained on. We also leverage the same
structure in tasks, but by employing vision foundation models
trained on large datasets, we are able
to operate on novel objects in novel environments.

%% file: 3-method.tex
\begin{figure*}
\insertW{1.0}{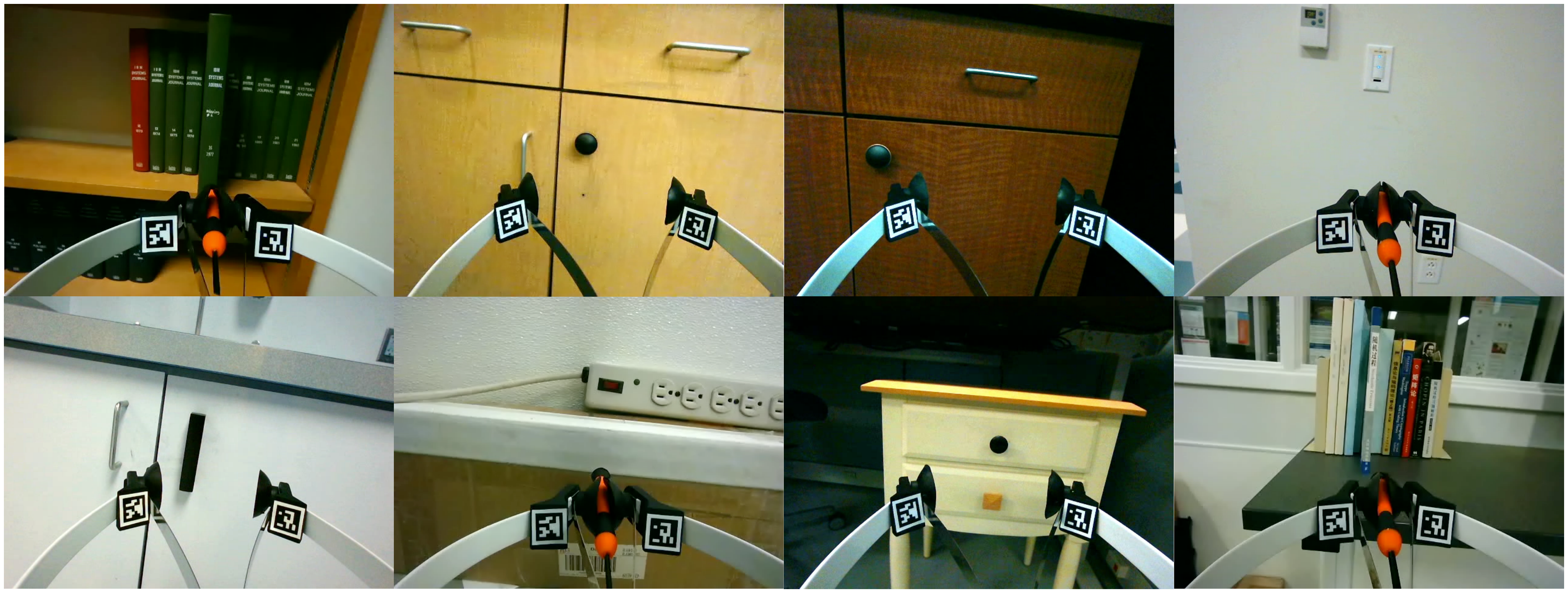}
\caption{\changedSinceFirstRAL{Our evaluation consists of 10 environments across 6 buildings, including 72 different object instances. Note that we exclusively test on \textit{novel} objects in \textit{novel} environments not used for training or development in any manner.}}
\figlabel{diversity}
\end{figure*}

\section{Task} Many everyday household tasks involve precise manipulation
followed by execution of a motion primitive. Examples include grasping a knob
or a handle to pull open drawers / cupboards, or pushing a button on a
microwave. 
\changedSinceFirstRAL{We consider two variants—interaction sites specified via a semantic label (e.g. knob or handle) or via a user-defined point (e.g. a click on an image specifying the button to push)—for a total of six tasks, and assume that the motion primitive is provided or easily specified.
Our goal is to enable a low DoF commodity
mobile manipulator equipped with a RGB-D wrist camera to accomplish such high precision tasks
in previously unseen environments.}

\section{Method}
\seclabel{method}

\changedSinceFirstRAL{At a high level, \name uses visual servoing on eye-in-hand images to guide the end-effector to the interaction site. One of the key innovations is leveraging vision foundation models to detect or track the target for visual feedback. Additionally, to handle occlusion from the end-effector, we first paint it out before utilizing the vision foundation models. Let $I_t$ be the wrist image and let $x_t$ be the robot configuration.}
The output actions are computed as follows:
\begin{equation}
a_t = \pi\left(x_t, g\left(f\left(I_t, [I_1, \ldots, I_{t-1}]\right)\right)\right),
\end{equation}
where $f(I, \bfI)$ is a video inpainting function that paints out the
end-effector from image $I$ using images in $\bfI$ as reference, $g(I)$
localizes the target in 3D in the wrist camera frame, and $\pi$ computes the
desired joint velocities using the current robot state $x_t$ and the current
target location output by $g$. \figref{overview} shows an overview of our
proposed method and we describe each component below.

\textbf{Inpainting.}
Given RGB images from the wrist camera, the inpainting function $f$ uses past
frames from the wrist camera to inpaint the current frame $I_t$. 
We utilize a video inpainting method (as opposed to an image inpainting method)
for better performance: a video inpainting model has access to previous frames
(where the object may be unoccluded), which can lead to improved
inpainting. 
\changedSinceFirstRAL{We adopt ProPainter~\cite{zhou2023propainter}, a transformer-based video inpainting model, to realize $f$.

If the target object is occluded in the first frame, even video inpainting may fail to reconstruct it. To address this, we execute a one-time “look-around” primitive that moves the end-effector vertically and laterally to gather scene context. We then use the resulting ten frames as input to the inpainting model, which is limited to processing the ten most recent images to reduce inference time.}

Out-painting the end-effector also requires a mask
of the end-effector.  We use a manually constructed mask that coarsely
covers the end-effector.  We find this to work better for out-painting than a
fine mask of the end-effector obtained using the image segmentation model
SAM~\cite{kirillov2023segment}.

\input{table_results.tex}

\textbf{Interaction Site Localization.}
Given an image with the end-effector painted out, our next goal is to localize
the object of interest to obtain the 3D location of the target. We handle the two
specifications for the interaction site, via a semantic label or a user click,
separately as described below.
\begin{itemize}[leftmargin=*, topsep=0pt, itemsep=0pt, parsep=0pt]
\item \textbf{Detection.}
For semantically specified targets (\eg knobs / handles), we use Detic
\cite{zhou2022detecting}, an open-vocabulary detector trained on large-scale
datasets. We prompt Detic with the object class `handle' for handles and `knob'
for knobs.
If Detic detects multiple handles in the image, we select the handle closest to
the center of the image. Detic also outputs a mask for the object.  We compute the
center of the mask and use this as 2D position of the object of interest.
\item \textbf{Tracking.}
For tasks specified via a user click, \eg the point on the book or the button in \figref{teaser}, we make use of CoTracker \cite{karaev2023cotracker}, a
point tracking method for videos. Given a point in the first frame, CoTracker
is able to track it over subsequent frames seen during execution. CoTracker
notes that tracking performance is better when tracking many points together.
We therefore sample 40 points randomly around the user click and find this to
drastically improve tracking performance.
\end{itemize}

\changedSinceFirstRAL{Both methods yield a 2D target location, which we lift to 3D using the depth image. 
If the target lies within the end-effector mask (i.e. is occluded), we use 
the depth from the nearest prior frame.}

\textbf{Closed-loop Control.}
Given the interaction sites' 3D location, we employ visual servoing to realize $\pi$ to compute
velocity control commands. Visual servoing computes \textit{whole-body} velocity 
control commands that minimize the distance between the end-effector 
and the 3D target point~\cite{chaumette2016visual}.

%% file: table_results.tex
\definecolor{Gray}{gray}{0.85}
\newcolumntype{g}{>{\columncolor{Gray}}c}
\renewcommand{\arraystretch}{1.2}
\begin{table*}
\centering
\caption{\changedSinceFirstRAL{{\bf Execution Success Rates.} We compare Servoing with Vision Models (\name) to a previous system (MOSART~\cite{gupta2024opening}), open-loop execution using targets computed in the wrist camera, and a version of \name without inpainting in 10 novel environments across 6 buildings. Tasks require precise control and open-loop execution fails. MOSART's contact correction works for handles but struggles with knobs and it cannot handle user-clicked targets. Inpainting matters for semantic targets.}}
\tablelabel{results}
\small
\begin{tabular}{lccccccg}
\toprule
                            
                            & \multicolumn{3}{c}{\bf Semantic Targets} & \multicolumn{3}{c}{\bf User-clicked Targets} & \multirow{2}{*}{\bf Total} \\
                            \cmidrule(lr){2-4} \cmidrule(l){5-7}
                            & \shortstack{\bf Knobs\\(Cabinets)} 
                            & \shortstack{\bf Knobs\\(Drawers)} 
                            & \shortstack{\bf Handles\\(Cabinets)} 
                            & \shortstack{\bf Light\\ \bf Button} 
                            & \shortstack{\bf Book\\ \bf Push} 
                            & \shortstack{\changedSinceFirstRAL{\bf Toggle}\\ \changedSinceFirstRAL{\bf Switch}} 
                            & \\
\midrule
MOSART~\cite{gupta2024opening}[i.e. Open Loop (Head Camera)]
                              & 1/7     & 0/7     & \bf 6/7 & -       & -       & -       & \changedSinceFirstRAL{7/42} \\
Open-Loop (Eye-in-Hand)       & 2/7     & 0/7     & \bf 6/7 & \changedSinceFirstRAL{0/7}     & \changedSinceFirstRAL{\bf 4/7} & \changedSinceFirstRAL{0/7}     & \changedSinceFirstRAL{12/42} \\
\name w/o inpainting           & \bf 6/7 & 3/7     & 4/7     & \bf \changedSinceFirstRAL{5/7} & \bf \changedSinceFirstRAL{4/7} & \bf \changedSinceFirstRAL{4/7} & \changedSinceFirstRAL{26/42} \\
\name (Ours)                          & \bf 6/7 & \bf 5/7 & \bf 6/7 & \bf \changedSinceFirstRAL{5/7} & \bf \changedSinceFirstRAL{4/7} & \bf \changedSinceFirstRAL{4/7}  & \bf \changedSinceFirstRAL{30/42} \\
\bottomrule
\end{tabular}
\normalsize
\end{table*}

%% file: 4-experiments.tex
\section{Experiments}
\seclabel{experiments}
\changedSinceFirstRAL{We conduct experiments in 10 previously unseen environments across 6 previously unseen buildings including 72 different object instances (see \figref{diversity}).}
Our experiments are designed to test a) the extent to which open-loop execution is an issue for precise mobile manipulation tasks, b) the effectiveness of blind proprioceptive correction techniques, c) whether object detectors and point trackers perform reliably in wrist camera images, d) the impact of end-effector occlusion and the efficacy of video inpainting in mitigating it, and e) how our proposed \name framework compares to large-scale imitation learning.

\subsection{Tasks and Experimental Setup}

\changedSinceFirstRAL{We use the Stretch RE2, a commodity mobile manipulator with a 5DOF arm on a non-holonomic base, upgraded with the Dex Wrist 3 and an eye-in-hand RGB-D camera (Intel D405).
We evaluate \name across 3 task families: a) grasping knobs to open drawers/cabinets, b) grasping handles to open cabinets, and c) pushing objects (buttons, books, switches).}
Our focus is on generalization. 
{\it
Therefore, we exclusively test on previously unseen instances, not used during
development.}

All tasks involve precise manipulation, followed by execution of a motion
primitive. {\bf For the pushing tasks}, the precise motion is to get the
end-effector exactly at the indicated point and the motion primitive is to push
in the direction perpendicular to the surface, before retracting the end-effector 
upon contact. The robot is positioned such
that the target position is within the field of view of the wrist camera. A user
selects the point of pushing via a mouse click on the wrist camera image. The
goal is to push at the indicated location. Success is determined by whether the
push results in the desired outcome (light turns on / off or book gets pushed in). 
\changedSinceFirstRAL{Since the rubber gripper bends on contact, we attach a rigid tool and account for its geometry during servoing.}

\changedSinceFirstRAL{{\bf For the opening articulated object tasks}, the precise manipulation involves grasping a knob or handle, followed by whole-body motion to open the cabinet. We adopt MOSART~\cite{gupta2024opening} for this motion. Each episode begins with the robot $\approx1.5m$ from the object, visible from the head camera. MOSART computes articulation parameters and navigates to a pre-grasp pose. \name (or baseline) then centers the gripper around the knob/handle, after which MOSART resumes: it extends until contact, closes the end-effector, and finally executes the predicted motion plan. Following \cite{gupta2024opening}, success is defined as opening the cabinet by over $60^\circ$ or the drawer by more than 24cm.}

For the precise manipulation part, all baselines consume the current and
previous RGB-D images from the wrist camera and output full body motor
commands.

\subsection{Baselines}
\changedSinceFirstRAL{We compare against four other methods for the precise manipulation part of
these tasks.}

\begin{figure*}[t]
\insertW{1.0}{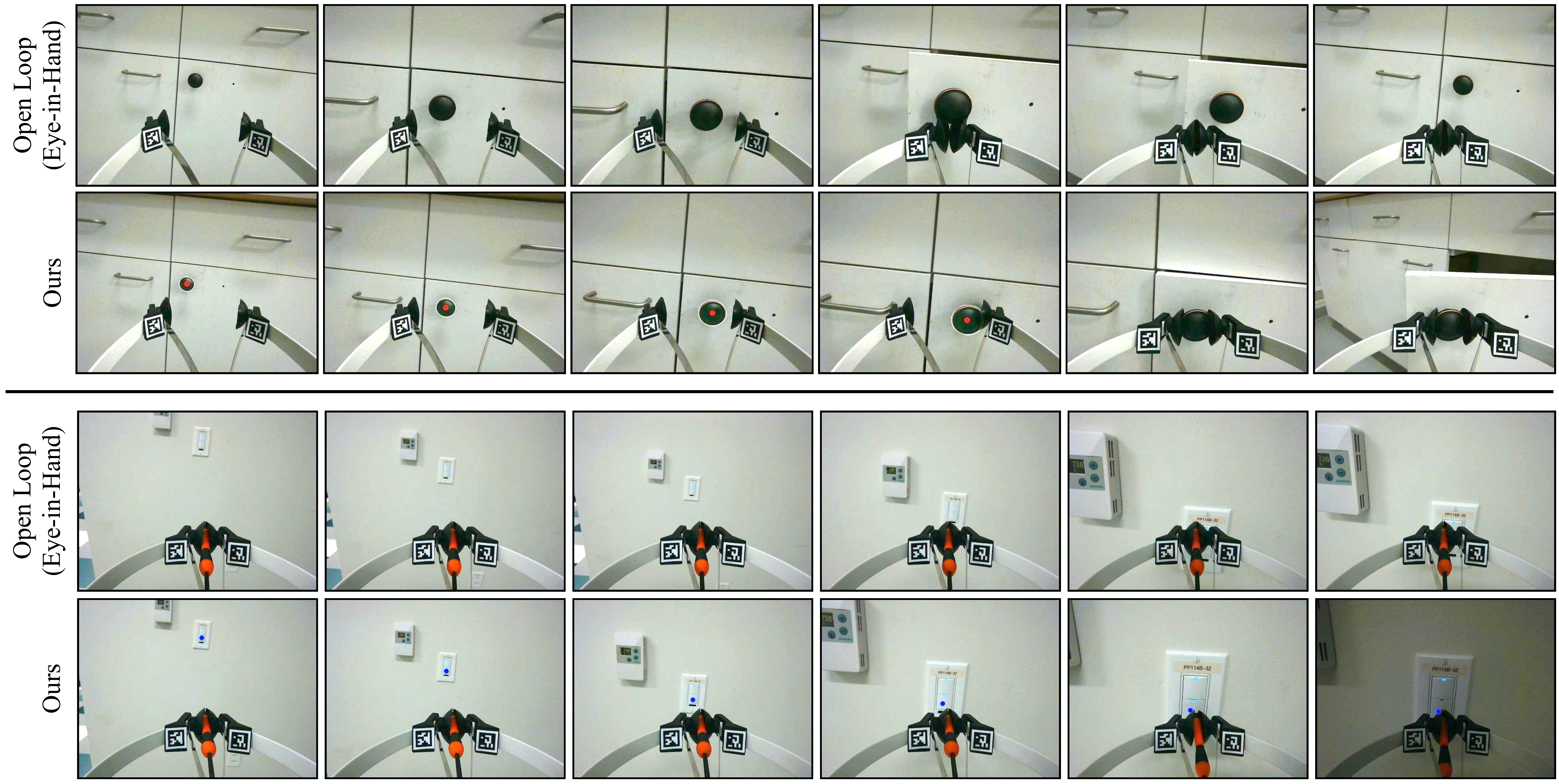}
\caption{\changedSinceFirstRAL{{\bf \name \vs Open-Loop (Eye-in-Hand) baseline.} \textbf{(top)} In opening a cabinet with a knob, slight errors in getting to the target cause the end-effector to slip off, leading to failure for the baseline, whereas our method is able to successfully complete the task. \textbf{(bottom)} Slight errors in getting to the target cause failure, whereas \name successfully turns the lights off. Note the high quality of CoTracker's track ({\color{blue} blue dot}).}}
\figlabel{rollout}
\end{figure*}

\textbf{MOSART~\cite{gupta2024opening}.}
The recent modular system for opening cabinets and drawers~\cite{gupta2024opening}
reports impressive performance with open-loop control (using the head camera from 1.5m away), combined with proprioception-based feedback to 
compensate for errors in perception and control when interacting with handles. 
We test if such correction is also sufficient for interacting with knobs. Note 
that such correction is not possible for the smaller buttons and pliable books.

\textbf{Open-Loop (Eye-in-Hand).} To assess the precision requirements of
the tasks and to set it in context with the manipulation capabilities of the
robot platform, this baseline uses open-loop execution starting from estimates
for the 3D target position from the first wrist camera image.

\textbf{\name (no inpainting).} 
\changedSinceFirstRAL{To assess the negative impact of the end-effector’s presence in the wrist camera image during
manipulation, we ablate the use of inpainting.}

\textbf{Robot Utility Models (RUM)~\cite{etukuru2024robot}.}
For the opening articulated object tasks, we also compare to Robot Utility Models (RUM), 
a closed-loop imitation learning method recently proposed by Etukuru et al. \cite{etukuru2024robot}.
RUM is trained on a substantial dataset comprising expert demonstrations, including 
1,200 instances of cabinet opening and 525 of drawer opening, gathered from roughly 
40 different environments.
This dataset stands as the most extensive imitation 
learning dataset for articulated object manipulation to date, establishing RUM as a 
strong baseline for our evaluation.

For RUM, similar to \name, we use MOSART to compute articulation
parameters and convey the robot to a pre-grasp location
with the target handle in view of the wrist camera.
One of RUM's assumptions is a good view of the handle.
To benefit RUM, we try out three different heights of the wrist camera,
and \textit{report the best result for RUM.}

\subsection{Results}

\changedSinceFirstRAL{\tableref{results} and \tableref{rum} present results from our experiments. 
We discuss our key
experimental findings from our large-scale evaluation spanning real world experiments across 10 novel environments below.}

\changedSinceFirstRAL{\textbf{Vision foundation models with servoing enable generalizable and precise mobile manipulation.}
Our approach \name successfully 
solves 71\% of 
unseen object instances in unseen
environments that were not used for development.} %

\textbf{Vision foundation models work reasonably well even on wrist camera images.}
Inpainting works well on wrist camera images (see \figref{occlusion}).
Closing the loop using feedback from vision detectors and point trackers on
wrist camera images also work well, particularly when we use inpainted images.
See some example detections and point tracks in \figref{rollout}. 
Detic~\cite{zhou2022detecting} was able to reliably detect the knobs and
handles, and CoTracker~\cite{karaev2023cotracker} was able to successfully track
the point of interaction letting us solve 30/42 task instances.

\textbf{Closing the loop is necessary for these precise tasks.} 
While the proprioception-based strategies proposed in MOSART~\cite{gupta2024opening}
work out for handles, they are inadequate for targets like knobs and just
do not work for tasks like pushing buttons. Using estimates from the wrist
camera is better, but open-loop execution still fails for knobs and pushing
buttons (see \figref{rollout}).

\textbf{Erroneous detections without inpainting hamper performance on 
handles and our end-effector out-painting strategy effectively mitigates it.} 
As shown in \figref{occlusion}, presence of the end-effector caused the object
detector to misfire, leading to failed execution. Our out-painting approach
mitigates this issue, leading to a higher success rate than the 
approach without out-painting. 
Interestingly, CoTracker~\cite{karaev2023cotracker} is quite robust
to occlusion and does not benefit
from inpainting.

\begin{table}[t]
\caption{Comparison of \name \vs RUM~\cite{etukuru2024robot}, a recent large-scale end-to-end imitation learning method trained on 1200 demos for opening cabinets and 525 demos for opening drawers across 40 different environments. \changedSinceFirstRAL{Our evaluation spans objects from 9 previously unseen environments across 6 buildings. \name significantly outperforms RUM. $^*$Note that RUM was not trained on knobs.}}
\setlength{\tabcolsep}{6pt} %
\centering
\resizebox{\linewidth}{!}{
\begin{tabular}{lccccg}
\toprule
& \multicolumn{2}{c}{\bf Handle} & \multicolumn{2}{c}{\bf Knobs} & \bf \multirow{2}{*}{\bf Total} \\
\cmidrule(lr){2-3} \cmidrule(lr){4-5}
& \bf Cabinets & \bf \changedSinceFirstRAL{Drawer} & \bf Cabinets & \bf Drawer & \\
\midrule
RUM~\cite{etukuru2024robot}   & \changedSinceFirstRAL{2/8} & \changedSinceFirstRAL{4/7} & \changedSinceFirstRAL{0$^*$/7} & \changedSinceFirstRAL{1$^*$/8} & \changedSinceFirstRAL{7/30} \\
\name (Ours)                  & \changedSinceFirstRAL{6/8} & \changedSinceFirstRAL{7/7} & \changedSinceFirstRAL{4/7} & \changedSinceFirstRAL{5/8} & \changedSinceFirstRAL{22/30} \\
\bottomrule
\end{tabular}}

\tablelabel{rum}
\end{table}

\begin{figure*}[t]
\insertW{1.0}{figures/rum_figure_smaller.pdf}
\caption{\changedSinceFirstRAL{{\bf \name vs RUM \cite{etukuru2024robot}}. We show roll-outs of both \name and RUM: at the start of the episode, at the start of the closed-loop grasping, just before closing the gripper, and during manipulation.}}
\figlabel{rum_comparison}
\end{figure*}

\begin{figure*}[t]
\insertW{1.0}{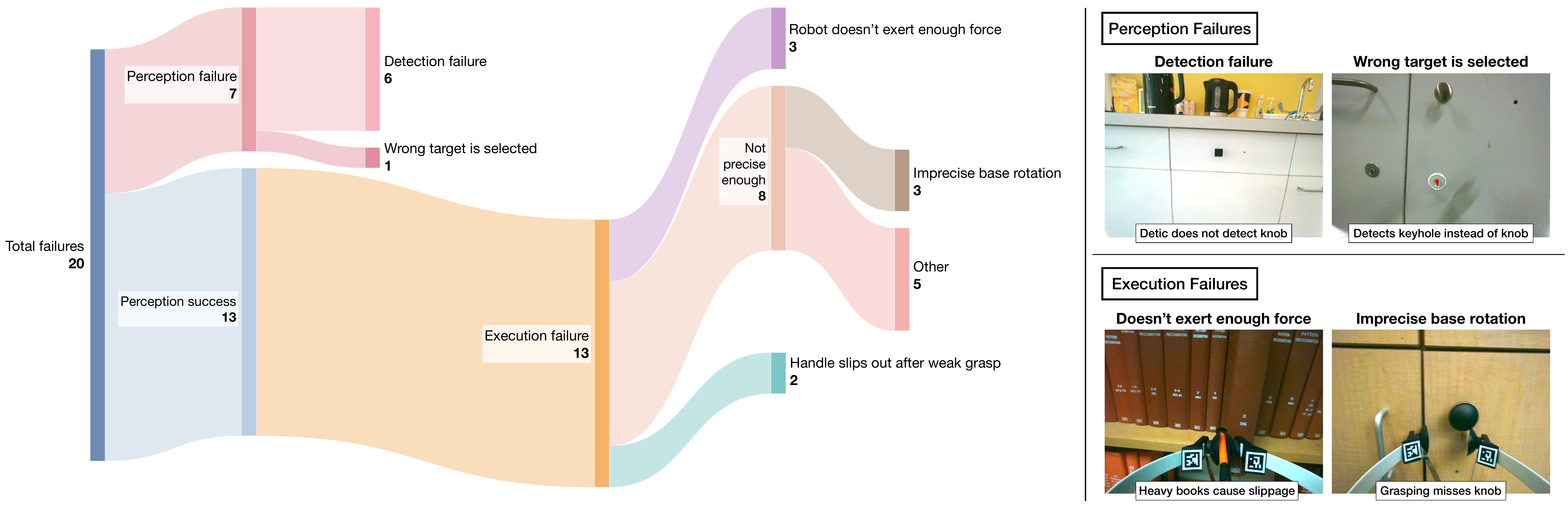}
\caption{\changedSinceFirstRAL{{\bf \name Failure Modes}. We characterize the failure modes of \name of all the 20 failures we encountered in \tableref{results} and \tableref{rum}. On the right, we show some examples of these failure modes.}}
\figlabel{waterfall_chart}
\end{figure*}

\changedSinceFirstRAL{\textbf{Closed-loop imitation learning struggles on novel objects.}
As presented in \tableref{rum}, \name significantly outperforms RUM in a paired evaluation on unseen objects across 9 novel environments spanning 6 buildings. For handles, RUM succeeds on 6/15 handles \vs SVM succeeds on 13/15 handles.
Although RUM was not explicitly trained on knobs, its exposure to a wide variety of handles during training suggested potential for generalization, which motivated our evaluation on knobs as well. While RUM achieves a non-zero success rate on knobs, it struggles relative to \name (see \figref{rum_comparison}).
One failure mode of RUM is failing to grasp the handle despite approaching it closely.}
These results demonstrate that a modular approach that leverages the broad generalization capabilities of vision foundation models is able to generalize much better than end-to-end imitation learning trained on 1000+ demonstrations, which must learn all aspects of the task from scratch.

\changedSinceFirstRAL{\textbf{\name Failure Modes.} 
Across 20 failures of \name in both \tableref{results} and \tableref{rum}, 35\% were perception errors and 65\% were execution errors (see \figref{waterfall_chart}). Within perception, mis-detections were surprisingly the dominant failure mode, despite using a vision foundation model. Within execution, insufficient precision was the dominant cause of failure. 
This suggests that future work should focus on strengthening eye-in-hand perception models and improving precise control under limited degrees of freedom.
}

%% file: 5-discussion.tex
\section{Discussion}
\seclabel{discussion}

In this paper, we describe \name, a framework for precise
manipulation tasks that involve reaching a precise interaction site followed by
execution of a primitive motion. Vision foundation models help us mitigate issues
caused by occlusion by the end-effector, thereby enabling the use of
off-the-shelf open-vocabulary object detectors and point trackers to estimate
servoing targets during execution. 
The use of strong off-the-shelf models also provides broad generalization for perception, while the use of servoing provides robust control.
\changedSinceFirstRAL{This enables \name to solve tasks in novel environments on novel objects,
obtaining a 71\% zero-shot success rate across 6 precise mobile manipulation
tasks.}
Perhaps most surprisingly, despite being modular, \name outperforms the strong end-to-end imitation learning system RUM~\cite{etukuru2024robot} which was trained on over 1000 demonstrations.
This is particularly 
striking as imitation learning is often the tool of choice for precise tasks requiring closed-loop execution.

\section{Limitations} Even though \name performs quite well across many
tasks on novel objects in novel environments, it suffers from some
shortcomings. Running these large vision models is computationally expensive
and we have to offload computation to a A40 GPU sitting in a server. Even with
this GPU, we are only able to run the vision pipeline at a 0.1 Hz
leading to slow executions. 
Building vision models specialized to wrist camera images may work better and
faster.
A second limitation is the reliance on depth from the wrist camera which may
be poor in some situations \eg shiny or dark objects. Use of learned
disparity estimators~\cite{xu2023iterative} with stereo images could
mitigate this. As our focus is on the precise reaching of interaction sites,
we work with a hand-crafted task decomposition into the interaction site and
primitive motion. In the future, we could obtain such a decomposition using
LLMs, or from demonstrations, thereby expanding the set of tasks we can tackle.

%% file: 6-acknowledgements.tex
\section{Acknowledgements}
\seclabel{ack}

This material is based upon work supported by DARPA (Machine Common Sense program), an NSF CAREER Award (IIS-2143873), and the Andrew T. Yang Research and Entrepreneurship Award. We are grateful to the Centre for Autonomy for lending us the Stretch RE2 robot used in this work. We thank Aditya Prakash and Xiaoyu Zhang for their feedback on manuscript.

%% file: root.bbl
\begin{thebibliography}{10}
\providecommand{\url}[1]{#1}
\csname url@samestyle\endcsname
\providecommand{\newblock}{\relax}
\providecommand{\bibinfo}[2]{#2}
\providecommand{\BIBentrySTDinterwordspacing}{\spaceskip=0pt\relax}
\providecommand{\BIBentryALTinterwordstretchfactor}{4}
\providecommand{\BIBentryALTinterwordspacing}{\spaceskip=\fontdimen2\font plus
\BIBentryALTinterwordstretchfactor\fontdimen3\font minus
  \fontdimen4\font\relax}
\providecommand{\BIBforeignlanguage}[2]{{%
\expandafter\ifx\csname l@#1\endcsname\relax
\typeout{** WARNING: IEEEtran.bst: No hyphenation pattern has been}%
\typeout{** loaded for the language `#1'. Using the pattern for}%
\typeout{** the default language instead.}%
\else
\language=\csname l@#1\endcsname
\fi
#2}}
\providecommand{\BIBdecl}{\relax}
\BIBdecl

\bibitem{chaumette2016visual}
F.~Chaumette, S.~Hutchinson, and P.~Corke, ``Visual servoing,'' \emph{Springer
  handbook of robotics}, pp. 841--866, 2016.

\bibitem{zhou2022detecting}
X.~Zhou \emph{et~al.}, ``Detecting twenty-thousand classes using image-level
  supervision,'' in \emph{ECCV}, 2022.

\bibitem{karaev2023cotracker}
N.~Karaev \emph{et~al.}, ``{CoTracker}: It is better to track together,'' in
  \emph{ECCV}, 2024.

\bibitem{chang2023look}
M.~Chang, A.~Prakash, and S.~Gupta, ``Look ma, no hands! agent-environment
  factorization of egocentric videos,'' in \emph{NeurIPS}, 2023.

\bibitem{zhou2023propainter}
S.~Zhou, C.~Li, K.~C. Chan, and C.~C. Loy, ``{ProPainter}: Improving
  propagation and transformer for video inpainting,'' in \emph{ICCV}, 2023.

\bibitem{lee2019copy}
S.~Lee, S.~W. Oh, D.~Won, and S.~J. Kim, ``Copy-and-paste networks for deep
  video inpainting,'' in \emph{ICCV}, 2019.

\bibitem{etukuru2024robot}
H.~Etukuru \emph{et~al.}, ``Robot utility models: General policies for
  zero-shot deployment in new environments,'' in \emph{ICRA}, 2025.

\bibitem{radford2021learning}
A.~Radford \emph{et~al.}, ``Learning transferable visual models from natural
  language supervision,'' in \emph{ICML}, 2021.

\bibitem{achiam2023gpt}
J.~Achiam, S.~Adler, S.~Agarwal, L.~Ahmad, I.~Akkaya, F.~L. Aleman, D.~Almeida,
  J.~Altenschmidt, S.~Altman, S.~Anadkat \emph{et~al.}, ``{GPT-4} technical
  report,'' \emph{arXiv preprint arXiv:2303.08774}, 2023.

\bibitem{kirillov2023segment}
A.~Kirillov, E.~Mintun, N.~Ravi, H.~Mao, C.~Rolland, L.~Gustafson, T.~Xiao,
  S.~Whitehead, A.~C. Berg, W.-Y. Lo, P.~Dollár, and R.~Girshick, ``Segment
  anything,'' in \emph{ICCV}, 2023.

\bibitem{dosovitskiy2021image}
A.~Dosovitskiy \emph{et~al.}, ``An image is worth 16x16 words: Transformers for
  image recognition at scale,'' in \emph{ICLR}, 2021.

\bibitem{zareian2021open}
A.~Zareian, K.~D. Rosa, D.~H. Hu, and S.-F. Chang, ``Open-vocabulary object
  detection using captions,'' in \emph{CVPR}, 2021.

\bibitem{sohl2015deep}
J.~Sohl-Dickstein, E.~Weiss, N.~Maheswaranathan, and S.~Ganguli, ``Deep
  unsupervised learning using nonequilibrium thermodynamics,'' in \emph{ICML},
  2015.

\bibitem{rombach2021high}
R.~Rombach \emph{et~al.}, ``High-resolution image synthesis with latent
  diffusion models,'' in \emph{CVPR}, 2022.

\bibitem{ho2020denoising}
J.~Ho, A.~Jain, and P.~Abbeel, ``Denoising diffusion probabilistic models,''
  \emph{NeurIPS}, 2020.

\bibitem{song2020score}
Y.~Song, J.~Sohl-Dickstein, D.~P. Kingma, A.~Kumar, S.~Ermon, and B.~Poole,
  ``Score-based generative modeling through stochastic differential
  equations,'' \emph{arXiv preprint arXiv:2011.13456}, 2020.

\bibitem{bahl2022human}
S.~Bahl \emph{et~al.}, ``Human-to-robot imitation in the wild,'' in \emph{RSS},
  2022.

\bibitem{harley2022particle}
A.~W. Harley, Z.~Fang, and K.~Fragkiadaki, ``Particle video revisited: Tracking
  through occlusions using point trajectories,'' in \emph{ECCV}, 2022.

\bibitem{doersch2023tapir}
C.~Doersch, Y.~Yang, M.~Vecerik, D.~Gokay, A.~Gupta, Y.~Aytar, J.~Carreira, and
  A.~Zisserman, ``Tapir: Tracking any point with per-frame initialization and
  temporal refinement,'' in \emph{ICCV}, 2023.

\bibitem{shi1994good}
J.~Shi and C.~Tomasi, ``Good features to track,'' in \emph{CVPR}, 1994.

\bibitem{wen2023any}
C.~Wen, X.~Lin, J.~So, K.~Chen, Q.~Dou, Y.~Gao, and P.~Abbeel, ``Any-point
  trajectory modeling for policy learning,'' in \emph{RSS}, 2024.

\bibitem{bharadhwaj2024track2act}
H.~Bharadhwaj, R.~Mottaghi, A.~Gupta, and S.~Tulsiani, ``Track2act: Predicting
  point tracks from internet videos enables diverse zero-shot robot
  manipulation,'' in \emph{ECCV}, 2024.

\bibitem{corke1993visual}
P.~I. Corke, ``Visual control of robot manipulators--a review,'' \emph{Visual
  Servoing: Real-Time Control of Robot Manipulators Based on Visual Sensory
  Feedback}, pp. 1--31, 1993.

\bibitem{chaumette2006visual}
F.~Chaumette and S.~Hutchinson, ``{Visual servo control, Part I: Basic
  approaches},'' 2006.

\bibitem{collewet2011photometric}
C.~Collewet and E.~Marchand, ``Photometric visual servoing,'' \emph{TOR}, 2011.

\bibitem{bateux2016histograms}
Q.~Bateux and E.~Marchand, ``Histograms-based visual servoing,'' \emph{RA-L},
  2016.

\bibitem{lee2017learning}
A.~X. Lee, S.~Levine, and P.~Abbeel, ``Learning visual servoing with deep
  features and fitted q-iteration,'' in \emph{ICLR}, 2017.

\bibitem{bateux2018training}
Q.~Bateux, E.~Marchand, J.~Leitner, F.~Chaumette, and P.~Corke, ``Training deep
  neural networks for visual servoing,'' in \emph{ICRA}, 2018.

\bibitem{sadeghi2018sim2real}
F.~Sadeghi, A.~Toshev, E.~Jang, and S.~Levine, ``Sim2real viewpoint invariant
  visual servoing by recurrent control,'' in \emph{CVPR}, 2018.

\bibitem{sadeghi2019divis}
F.~Sadeghi, ``Divis: Domain invariant visual servoing for collision-free goal
  reaching,'' \emph{arXiv preprint arXiv:1902.05947}, 2019.

\bibitem{li2021survey}
C.~Li, B.~Li, R.~Wang, and X.~Zhang, ``A survey on visual servoing for wheeled
  mobile robots,'' \emph{International Journal of Intelligent Robotics and
  Applications}, vol.~5, no.~2, pp. 203--218, 2021.

\bibitem{qureshi2021rtvs}
M.~N. Qureshi \emph{et~al.}, ``Rtvs: A lightweight differentiable mpc framework
  for real-time visual servoing,'' in \emph{IROS}.\hskip 1em plus 0.5em minus
  0.4em\relax IEEE, 2021.

\bibitem{argus2020flowcontrol}
M.~Argus, L.~Hermann, J.~Long, and T.~Brox, ``Flowcontrol: Optical flow based
  visual servoing,'' in \emph{IROS}.\hskip 1em plus 0.5em minus 0.4em\relax
  IEEE, 2020.

\bibitem{crombez2021subsequent}
N.~Crombez, J.~Buisson, Z.~Yan, and Y.~Ruichek, ``Subsequent keyframe
  generation for visual servoing,'' in \emph{ICRA}, 2021.

\bibitem{tang2023industreal}
B.~Tang, M.~A. Lin, I.~Akinola, A.~Handa, G.~S. Sukhatme, F.~Ramos, D.~Fox, and
  Y.~Narang, ``Industreal: Transferring contact-rich assembly tasks from
  simulation to reality,'' in \emph{RSS}, 2023.

\bibitem{luo2025precise}
J.~Luo, C.~Xu, J.~Wu, and S.~Levine, ``Precise and dexterous robotic
  manipulation via human-in-the-loop reinforcement learning,'' \emph{Science
  Robotics}, vol.~10, Aug. 2025.

\bibitem{goyal2024rvt}
A.~Goyal, V.~Blukis, J.~Xu, Y.~Guo, Y.-W. Chao, and D.~Fox, ``Rvt2: Learning
  precise manipulation from few demonstrations,'' \emph{RSS}, 2024.

\bibitem{collins2024forcesight}
J.~A. Collins, C.~Houff, Y.~L. Tan, and C.~C. Kemp, ``Forcesight: Text-guided
  mobile manipulation with visual-force goals,'' in \emph{ICRA}, 2024.

\bibitem{gupta2024opening}
A.~Gupta, M.~Zhang, R.~Sathua, and S.~Gupta, ``Demonstrating mosart: Opening
  articulated structures in the real world,'' \emph{RSS}, 2025.

\bibitem{pomerleau1991efficient}
D.~A. Pomerleau, ``Efficient training of artificial neural networks for
  autonomous navigation,'' \emph{Neural computation}, 1991.

\bibitem{schaal1996learning}
S.~Schaal, ``Learning from demonstration,'' in \emph{NeurIPS}, vol.~9, 1996.

\bibitem{chi2023diffusionpolicy}
C.~Chi \emph{et~al.}, ``Diffusion policy: Visuomotor policy learning via action
  diffusion,'' in \emph{RSS}, 2023.

\bibitem{mahi2022behavior}
N.~M. Mahi~Shafiullah, Z.~J. Cui, A.~Altanzaya, and L.~Pinto, ``Behavior
  transformers: Cloning $ k $ modes with one stone,'' \emph{arXiv e-prints},
  pp. arXiv--2206, 2022.

\bibitem{zhao2023learning}
T.~Zhao, V.~Kumar, S.~Levine, and C.~Finn, ``Learning fine-grained bimanual
  manipulation with low-cost hardware,'' in \emph{RSS}, 2023.

\bibitem{zhu2022viola}
Y.~Zhu, A.~Joshi, P.~Stone, and Y.~Zhu, ``Viola: Imitation learning for
  vision-based manipulation with object proposal priors,'' \emph{CoRL}, 2022.

\bibitem{khazatsky2024droid}
A.~Khazatsky, K.~Pertsch, S.~Nair, A.~Balakrishna, S.~Dasari, S.~Karamcheti,
  S.~Nasiriany, M.~K. Srirama, L.~Y. Chen, K.~Ellis \emph{et~al.}, ``Droid: A
  large-scale in-the-wild robot manipulation dataset,'' \emph{arXiv preprint
  arXiv:2403.12945}, 2024.

\bibitem{chi2024universal}
C.~Chi, Z.~Xu, C.~Pan, E.~Cousineau, B.~Burchfiel, S.~Feng, R.~Tedrake, and
  S.~Song, ``Universal manipulation interface: In-the-wild robot teaching
  without in-the-wild robots,'' in \emph{RSS}, 2024.

\bibitem{shafiullah2023bringing}
N.~M.~M. Shafiullah \emph{et~al.}, ``On bringing robots home,'' \emph{arXiv
  preprint arXiv:2311.16098}, 2023.

\bibitem{valassakis2022demonstrate}
E.~Valassakis, G.~Papagiannis, N.~Di~Palo, and E.~Johns, ``Demonstrate once,
  imitate immediately (dome): Learning visual servoing for one-shot imitation
  learning,'' in \emph{IROS}, 2022.

\bibitem{wen2022you}
B.~Wen, W.~Lian, K.~Bekris, and S.~Schaal, ``You only demonstrate once:
  Category-level manipulation from single visual demonstration,'' in
  \emph{RSS}, 2022.

\bibitem{vecerik2023robotap}
M.~Vecerik \emph{et~al.}, ``{RoboTAP}: Tracking arbitrary points for few-shot
  visual imitation,'' \emph{ICRA}, pp. 5397--5403, 2024.

\bibitem{xu2023iterative}
G.~Xu, X.~Wang, X.~Ding, and X.~Yang, ``Iterative geometry encoding volume for
  stereo matching,'' in \emph{CVPR}, 2023.

\end{thebibliography}
